\documentclass{isprs} 
\usepackage{subfigure}
\usepackage{setspace}
\usepackage{geometry} 
\usepackage{epstopdf}
\usepackage{verbatim}
\usepackage[labelsep=period]{caption}  
\usepackage[british]{babel} 
\usepackage[hang]{footmisc}
\usepackage{color}
\usepackage{colortbl}
\usepackage{amsmath,amsfonts}
\usepackage{comment}

\begin{document}

\title{Depth Supervised Neural Surface Reconstruction from Airborne Imagery}
\date{}

\author{ V. Hackstein\textsuperscript{1}\thanks{Corresponding author}\enspace, P. Fauth-Mayer\textsuperscript{1}, M. Rothermel\textsuperscript{1}, N. Haala\textsuperscript{2}}


\address{	\textsuperscript{1 }nFrames ESRI, Germany - (vhackstein, pfauthmayer, mrothermel)@esri.com\\ \textsuperscript{2 }Institute for Photogrammetry and Geoinformatics, University of Stuttgart, Germany -  norbert.haala@ifp.uni-stuttgart.de}



\abstract{While originally developed for novel view synthesis, Neural Radiance Fields (NeRFs)  have recently emerged as an alternative to multi-view stereo (MVS). Triggered by a manifold of research activities, promising results have been gained  especially for texture-less, transparent, and reflecting surfaces, while such scenarios remain challenging for traditional MVS-based approaches. However, most of these investigations focus on close-range scenarios, with studies for airborne scenarios still missing. For this task, NeRFs face potential difficulties at areas of low image redundancy and weak data evidence, as often found in street canyons, facades or building shadows. Furthermore, training such networks is computationally expensive. Thus, the aim of our work is twofold: First, we investigate the applicability of NeRFs for aerial image blocks representing different characteristics like nadir-only, oblique and high-resolution imagery. Second, during these investigations we demonstrate the benefit of integrating depth priors from tie-point measures, which are provided during presupposed Bundle Block Adjustment. Our work is based on the state-of-the-art framework VolSDF, which models 3D scenes by signed distance functions (SDFs), since this is more applicable for surface reconstruction compared to the standard volumetric representation in vanilla NeRFs. For evaluation, the NeRF-based reconstructions are compared to results of a publicly available benchmark dataset for airborne images.
}


\maketitle


\section{Introduction}\label{Introduction}


Image based 3D surface reconstruction is useful for many applications including urban modeling, environmental studies, simulations, robotics, and virtual reality. Typically, this task is solved by matured photogrammetric pipelines. As a first step, a Structure-from-Motion (SfM) approach estimates camera poses for further processing. While this step already provides a sparse reconstruction from tie point measurement as required by the Bundle Block Adjustment, dense 3D point clouds or meshes are generated by a multi-view stereo (MVS) pipeline in the second step. Examples of state-of-the-art approaches developed during the last decade are COLMAP ~\cite{SchoenbergerMVS}, PMVS ~\cite{Furu:2010:PMVS} or SURE ~\cite{Rothermel2012SURE}. Despite the considerable reconstruction quality available from such pipelines, they still suffer from problems at fine geometric structures, texture-less regions, and especially at non-Lambertian surfaces, i.e. at semi-transparent objects or reflections. Also due to these remaining issues, alternative approaches  based on Neural Radiance Fields (NeRF) gained considerable attention. Originally, this technique was developed for synthesizing novel views of complex scenes by  using a sparse set of input views~\cite{Mildenhall2022}. For this purpose, a neural network provides an implicit representation of the surface geometry as well as the appearance of the scene. This representation is then used to generate synthetic views by neural volume rendering. The neural network is trained to minimize the difference between the observed images and the corresponding virtual view of the scene. While NeRFs were originally motivated by visualisation applications, a considerable part of research work meanwhile focuses on 3D reconstruction ~\cite{li2023neuralangelo,neus2}. However, the corresponding experiments are typically limited to reconstructions of close range scenes while investigations using aerial imagery are just emerging ~\cite{xu2024multitiling}. Such aerial applications make specific demands like the timely processing of large areas including the reconstruction at different scales, or challenging scenarios like street canyons, glass facades or shadowed areas. Furthermore, despite impressive results presented so far, NeRF-based surface reconstruction frequently suffers from challenges for low image redundancy and weak data evidence, while the computational effort for training the neural network is still considerable. To mitigate this problem recent works use additional cues for initialization or supervision during training. One option to support dense reconstruction is to integrate \textit{a priori} structural information as provided from the sparse SfM point cloud to this process. Since the volumetric representation of vanilla NeRFs is sub-optimal for surface reconstruction tasks, approaches as \cite{VolSDF,wangNeuSLearningNeural2023}  model 3D scenes using signed distance functions (SDFs). By these means, VolSDF combines the advantages of volume rendering methods during training and implicit surface modeling for geometric reconstruction. As our main contribution, we integrate tie point supervision into VolSdf and evaluate its reconstruction capabilities for typical aerial nadir and oblique image blocks. 

The remainder of our paper is as follows: Section \ref{RelatedWork} gives a brief overview on classical MVS and the state-of-the-art on NeRF based reconstruction. Section \ref{Methodology} then presents our approach, which modifies the framework VolSDF \cite{VolSDF} to supervise and thus support the training process using SfM tie points. As discussed in section \ref{RelatedWork} this framework provides an easy accessible representation of 3D model geometry, which is well suited for regularization during training.
Section \ref{Evaluation} evaluates our pipeline for three aerial image sets featuring different configurations. These investigations on data typically used in professional aerial mapping are interesting from a practical point of view while investigating specific challenges of NeRF-based surface reconstruction. Such aerial image collections feature limited viewing directions and potentially suffer from restricted surface observation due to occlusions. Additional challenges are variances in lighting conditions and moving objects or shadows. We show that training by  tie-points supervision is crucial for fast convergence and mitigates convergence to local minima during training. This holds in particular true for demanding scenes featuring vegetation or contradictory data e.g moving shadows. For evaluation, the results of our NeRF based reconstruction are analyzed for three data sets, including a comparison to results of a benchmark on high density aerial image matching \cite{Haala2013TheLO}.

\section{Related Work}\label{RelatedWork}
\subsection{Classical MVS}
Taking a collection of images and their pixel-accurate poses as input, most prominent MVS systems reconstruct dense geometry in form of points or depth maps. Many approaches use stereo or multi-view stereo algorithms to reconstruct depth maps \cite{Galliani_2015_ICCV,SchoenbergerMVS,Rothermel2012SURE}. Another prominent line of work starts with a sparse set of points which are iteratively refined and densified \cite{Furu:2010:PMVS,goesele2007multi}. In a second step a globally consistent, topologically valid surface is extracted using volumetric reconstruction such as \cite{kazhdan2013screened,labatut2009robust,jancosek2011multi,fuhrmann2014floating,ummenhofer2015global}. To enhance detail, meshes can be further refined such that photoconsistency across views is maximized \cite{vu2012high,delaunoy2008minimizing}. Such reconstruction pipelines rely on a sequence of computational expensive optimization algorithms. Moreover, each module has to be carefully tuned with regard to its parameters and quality of input from upstream modules.       
\subsection{Neural Implicit Representations} 
\sloppy
The seminal work introducing NeRF \cite{mildenhall2020nerf} opened a new research path in the area of novel view synthesis. \cite{bruallaNeRFWildNeural2021} robustify vanilla NeRF for imagery with varying illumination conditions. \cite{barron2021mipnerf} show improved rendering quality by employing a training regime mitigating aliasing and accounting for the fact that pixels capture the scene with different ground resolution. Scalability for larger scenes is addressed in \cite{tancik2022blocknerf,xiangli2022bungeenerf,Turki_2022_CVPR}. \cite{liINGeoAcceleratingInstant2022,yu_and_fridovichkeil2021plenoxels,hu2023trimiprf} greatly improve training and inference times, by fully or partly replacing the original representation of scene geometry by spatial data structures such as multi-scale hash encoding or 3D MipMaps. 

NeRF style approaches target  the task of novel view-synthesis. They implicitly represent 3D geometry as density and the light emitted at a specific position in space, which impedes straight-forward surface regularization. Instead, methods targeting 3D reconstruction model the geometry by an implicit surface such as occupancy or signed distance functions. This enables the formulation of surface regularization losses and defines a global threshold required to extract surfaces using marching cubes \cite{marchingc}. Early work employed surface rendering \cite{DBLP:journals/corr/abs-1912-07372,yariv2020multiview}. However, geometry and radiance are only optimized near the surface hampering fast convergence for complex scenes. In contrast \cite{oechsleUNISURFUnifyingNeural2021,wangNeuSLearningNeural2023,VolSDF} implement volume rendering, optimizing geometry and radiance for an extended scene volume eventually approaching surface vicinity. This stabilizes convergence. Training and inference times can be considerably improved by efficient GPU implementation and incorporating multi scale hash encoding \cite{li2023neuralangelo,neus2}. Inspired by traditional MVS approaches, \cite{fuGeoNeusGeometryConsistentNeural2022,DBLP:journals/corr/abs-2112-09648} introduce losses encouraging multi-view photometric consistency of surface patches. Contrary to neural surface representations Gaussian splatting ~\cite{3DGaussianSplatting} represents the scene by splats (3D points, 3D covariances, color and opaqueness). Differentiable rendering of splats can be efficiently implemented on GPUs which allows for impressive rendering times. To reconstruct surfaces from splats \cite{guedon2023sugar} regularize 3D Gaussians and extract meshes by subsequent Poisson reconstruction.

Training of neural implicit representations is challenging for image collections featuring limited surface observations and challenging appearance.  Additional depth cues from monocular depth \cite{yuMonoSDFExploringMonocular2022} or RGBD-sensors \cite{azinovicNeuralRGBDSurface2022} can mitigate this problem. Similar to \cite{dengDepthsupervisedNeRFFewer2022} we supervise our reconstructions with SfM tie points. In the domain of remote sensing \cite{mariSatNeRFLearningMultiView2022,satmesh} train NeRFs or neural implicit surfaces for satellite imagery. \cite{Turki_2022_CVPR} propose a NeRF variant for large aerial scenes but focus on novel view synthesis. Most similar to our work, \cite{xu2024multitiling} provide an performance and quality evaluation of a scaleable MipNerf \cite{barron2021mipnerf} for aerial images. In contrast in this work we investigate  implicit neural surface reconstruction from aerial imagery.

\section{Methodology}\label{Methodology}
In this section we first review VolSDF \cite{VolSDF} as it is the base method for our implementation. We then explain our extension for depth supervision and implemented training schemes.
\subsection{Recap of VolSDF}
\paragraph{NeRF.} A neural radiance field is a neural function that takes spatial coordinates $\mathbf{x}$ and a viewing direction vector $\mathbf{v}$ as input which is mapped to a scalar density $\sigma$ and a color vector $\mathbf{c}$. 
$$F_{\Theta} : (\mathbf{x}, \mathbf{v}) 
\rightarrow ( \sigma, \mathbf{c}).$$
The function is modeled by two fully connected multilayer perceptrons (MLPs) encoding geometry and appearance respectively. The color $\mathbf{\hat{C}}$ of a pixel in an arbitrary view encoded by $F_{\Theta}$ can be composed using volume rendering. Let $\mathbf{r}$ (with direction $\mathbf{v}$) be the ray defined by the center of projection of a view and the pixel coordinate. We sample $N$ points $x_i, i \in [0,N]$ along $\mathbf{r}$, the distances between point samples are given by $\delta_i, i \in [0,N-1]$.  Using the quadrature based on the rectangle rule \cite{468400}, the discrete formulation of volume rendering is given by  
\begin{equation}\label{equ:volrendering}
\mathbf{\hat{C}}(\mathbf{r})=\sum_{i=1}^{N} T_i o_i \mathbf{c}_i.
\end{equation}
Thereby
\begin{equation}
o_i = 1 - \text{exp} \left( -\sigma_i \delta_i\right)
\end{equation}
is a notion of the emitted light or opacity.  
\begin{equation}
T_i = \text{exp} \left(-\sum_{j=1}^{i-1} \sigma_j \delta_j \right)
\end{equation}
represents the accumulated transparency along the ray up to the current position $r_i$. Thus, light emitted for samples $j<i$ reduces the contribution of colored light emitted for sample $i$ in the rendering equation \ref{equ:volrendering}. 

Since equation \ref{equ:volrendering} is differentiable, a loss minimizing photo-consistency across all views can be specified by the projection error:
\begin{equation}
\mathcal{L}_{RGB} = \| \mathbf{\hat{C}}(\mathbf{r}) - \mathbf{C} \|_2 .
\end{equation}

\paragraph{VolSDF - SDF representation.} In contrast to NeRF and variants, VolSDF models geometry by a signed distance function which only in a subsequent step is mapped to density. With $\Omega$ the 3D space occupied by some object and $\mathcal{M}$ the object boundary, the signed distance function can be formally written by 
\begin{equation}
	d_{\Omega} = (-1)^{\mathbf{1}_{\Omega}(\mathbf{x})}
\underset{\mathbf{y} \in \mathcal{M}}{\,\text{min}} \Vert \mathbf{x} - \mathbf{y} \Vert
\end{equation}
with 
\begin{equation}
\mathbf{1}_{\Omega}(\mathbf{x}) = 
			\begin{cases}
				1 \text{ if $\mathbf{x} \in \Omega$} \\
				0 \text{ if $\mathbf{x} \notin \Omega$}
			\end{cases} \text{.}
\end{equation}

To be able to employ volume rendering the signed distance is mapped to density with two learnable parameters $\beta$ and $\alpha$
\begin{equation} \label{eq:density}
	\sigma \left(\mathbf{x}\right) = \alpha \Psi_{\beta}\left(-d_{\Omega}\left(\mathbf{x}\right)\right)
\end{equation}
with
\begin{equation} \label{laplace}
	\Psi_{\beta}\left( s \right) = 
		\begin{cases}
		\frac{1}{2} \exp\left(\frac{s}{\beta}\right) & \text{if } s \leq 0 \\
		1 - \frac{1}{2} \exp\left(-\frac{s}{\beta}\right) & \text{if } s > 0.
		\end{cases}
\end{equation}

Surface points $\mathbf{x} \in \mathcal{M}$ have a constant density of $\frac{1}{2}\alpha$. Density smoothly decreases for increased distances from the surface.
This smoothness is controlled by $\beta$.
As $\beta$ approaches 0, $\Psi$ will converge to a step function, and $\sigma$ will converge to a scaled indicator function that maps all points inside the object to $\alpha$, and all other points to 0. More intuitively,
$\beta$ can be interpreted as a parameter encoding the confidence in the SDF in the current training stage. When the confidence is low, or equivalent, $\beta$ is high, points distant from the surface will map to larger densities and contribute to optimization. In later stages when the confidence is higher, only points close to the surface will contribute to optimization. Similar to \cite{VolSDF}, $\alpha$ is set to $\frac{1}{\beta}$ in all our experiments.

\paragraph{VolSDF - Regularization.} The SDF representation allows for regularization of the surface to cope with weak or contradictory image data. Similar to \cite{icml2020_2086} we use the eikonal loss
\begin{equation} \label{eq:eikonal_loss}
	\mathcal{L}_{eik} = (\Vert \nabla d(\mathbf{x}) \Vert - 1)^2.
\end{equation}
to enforce smoothness of the signed distance field.
Additionally we found that including a prior enforcing consistency of normals \cite{oechsleUNISURFUnifyingNeural2021} within an increased local neighborhood $\Delta \mathbf{x}$ benefits reconstruction
\begin{equation} \label{eq:unisurf-smooth}
	\mathcal{L}_{surf} = \Big\Vert \mathbf{n}(\mathbf{x}) - \mathbf{n}(\mathbf{x} + \Delta\mathbf{x}) \Big\Vert_{2}.
\end{equation}
The normal $\mathbf{n}$ is the gradient of the signed distance field and can be computed using double backpropagation \cite{OechsleVolumRendering}.

\paragraph{VolSDF - Sampling.} As in all NeRF-style methods a sampling strategy, ideally sampling points close to the correct surface but at the same time being able to recover/converge from inaccurate states of the NeRF is crucial. VolSDF ultimately places samples based on \textit{inverse transform sampling} of the discrete opacity function $o(i)$. The accuracy of $o(i)$ is influenced by the spatial extent of the ray where the samples are placed. Furthermore, for low sampling densities an approximation error is introduced by quadrature. VolSDF implements an iterative sampling mechanism which (a) bounds the approximation error of $o(i)$ and (b) adapts the sampling extent being closer to the surface with increased confidence of the SDF estimation. For details the reader is referred to the original publication. 

\
\subsection{Tie Point Supervision}

The input of VolSDF are poses which are computed within SfM  or aerial triangulation. As a side product $j$ tie points, each encoding homologous 2D image locations $\mathbf{x}_{i,j}$ across $i$ images and their corresponding 3D point $\mathbf{X}_j$, are generated. For each $\mathbf{x}_{i,j}$ a depth $d_{i,j}$ can be computed by projection. Similarly to \cite{dengDepthsupervisedNeRFFewer2022} we use tie points to initialize and supervise the training of VolSDF. More specifically, we follow \cite{azinovicNeuralRGBDSurface2022} and
sample two set of depths $\mathcal{S}^{tr}$ and $\mathcal{S}^{fs}$ along rays induced by $\mathbf{x}_{i,j}$.
$\mathcal{S}^{tr}$ contains samples $d_s$ close to the surface, $|d_{i,j}-d_s|<tr$. 
$\mathcal{S}^{fs}$ contains samples between the camera center and the surface point $d_{i,j}$ with $d_s \in \{ 0, d_{i,j}-tr\}$. A first loss enforces the predicted SDF $\hat{d_s}$ to correspond with $d_s$
\begin{equation} \label{eq:sdloss}
	\mathcal{L}_{tr} = \frac{1}{\vert\mathcal{R}\vert} \sum_{r \in \mathcal{R}} \frac{1}{\vert\mathcal{S}^{tr}\vert} \sum_{s \in \mathcal{S}^{tr}} \left(d_s - \hat{d}_s\right)^2,
\end{equation}
where $\mathcal{R}$ is a set of randomly sampled rays over all input images per training batch.
\cite{azinovicNeuralRGBDSurface2022} encourage a constant SDF value $tr$ in freespace. In contrast we relax that constraint and define a loss which only enforces SDF values larger than $t_r$ 
\begin{equation} \label{eq:fsloss}
	\mathcal{L}_{fs} =\frac{1}{\vert\mathcal{R}\vert} \sum_{r \in \mathcal{R}}^{N} \frac{1}{\vert\mathcal{S}^{fs}\vert} \sum_{s \in \mathcal{S}^{fs}} ReLU^2\left(tr - \hat{d}_s\right).
\end{equation}

We note the $d_s$ is only an approximation of the signed distance and rigorously valid in front-to-parallel settings, however we do not introduce any bias on the zero level set. For all experiments we set $tr$ to 30 times the GSD to safely exceed the noise levels.

\subsection {Implementation and Training Details}
Our model builds on the VolSDF implementation \cite{Yu2022SDFStudio}. It is composed of learnable multi-resolution hash grid encoding \cite{mullerInstantNeuralGraphics2022} and two MLPs with two layers of 256 neurons each.  We set the leaf size of the grid to match  the GSD of each evaluated dataset.
Our final training loss consists of five terms:

\begin{equation} \label{eq:loss_rgbdepth}
 \begin{split}
\mathcal{L} = \mathcal{L}_{RGB} & + \lambda_{eik} \mathcal{L}_{eik} + \lambda_{surf} \mathcal{L}_{surf} \\
+ &\lambda_{fs} \mathcal{L}_{fs} + \lambda_{tr} \mathcal{L}_{tr},
\end{split}
\end{equation}

where $\lambda_{*}$ are hyperparameters controlling the contribution of the respective loss terms. Their values were found by grid search, are constant in all experiments and listed in table \ref{tab:training_parameters}.
The network is optimized using the Adam optimizer \cite{kingma2017adam} with a learning rate of $lr = 5\mathrm{e}{-4}$ and exponential decay with rate $0.1$. The batch size remains constant and is set to $4096\,$ rays per training iteration. Training and inferences were run on an AMD Ryzen 3960X 24-Core CPU and a Nvidia
RTX 3090.

\begin{table}[h]
	\centering
		\begin{tabular}{|l|c|c|c|}\hline
			Parameter & Symbol & \multicolumn{2}{c|}{Values} \\\hline
                && 1st stage & 2nd stage  \\\hline
                Eikonal factor & $\lambda_{eik} $ & 0 & 5e-4  \\
                Surface smooth. factor & $\lambda_{surf} $ & 1e-2 & 5e-3 \\
                Surface smooth. radius & $R_{surf}$ & 35\,GSD & 35\,GSD \\
                Free-space factor & $\lambda_{fs}$ & 10 & 10\\
                Signed-distance factor & $\lambda_{tr}$ & 60 & 60\\
                Initial $\beta$ & $\beta_0$ & 0.001 & 0.001 \\
   \hline
		\end{tabular}
	\caption{Training parameters.}
\label{tab:training_parameters}
\end{table}
The training is split into two stages: a fast geometric initialization with depth supervision and smoothness regularization only, and a second stage additionally activating photometric supervision. The duration of the first stage is 1\,k epochs for all experiments. We evaluate results after training for 30\,k and 100\,k epochs.

\section{Evaluation}\label{Evaluation}

\subsection{Datasets}
We qualitatively evaluate our method on three datasets. These include two image blocks captured by professional large-format cameras in nadir only (\textit{Frauenkirche}) and oblique (\textit{Domkirk}) configuration. Images of \textit{Frauenkirche} are part of a benchmark on high density aerial image matching \cite{Haala2013TheLO}. Furthermore, we run tests on precisely georeferenced, high-resolution UAV images provided within a recent benchmark \textit{Hessigheim 3D} \cite{KOLLE2021100001,HAALAHessigheim}. More details can be found in table \ref{tab:dataset_infos}. The image collections cover challenging urban scenes, including thin structures, low data evidence (e.g. limited views, occlusions), photometric inconsistency and ambiguity (e.g. moving objects, shadows, and vegetation). 
We generate tie points for each dataset using a commercial AT software \cite{reality}.

\begin{table*}[h]
	\centering
		\begin{tabular}{|l|c|c|c|c|c|}\hline
			Dataset &  GSD & Configuration & Images & Pixels & Dimensions \\\hline
			\textit{Frauenkirche} & $10\,$cm & Aerial Nadir & 35 & $86\,$MPix & $95^3\,$m$^3$ \\
			\textit{Domkirk} & $3\,$cm & Aerial Oblique & 226 & $1362\,$MPix & $42^3\,$m$^3$   \\
			\textit{Hessigheim 3D}  & $1.5\,$cm & UAV & 217 & $1500\,$MPix & $27^3\,$m$^3$\\\hline
		\end{tabular}
	\caption{Datasets used for evaluation.}
\label{tab:dataset_infos}
\end{table*}

\subsection{Qualitative Evaluation}
\begin{figure}[ht!]
\begin{center}
		\includegraphics[width=1.0\columnwidth]{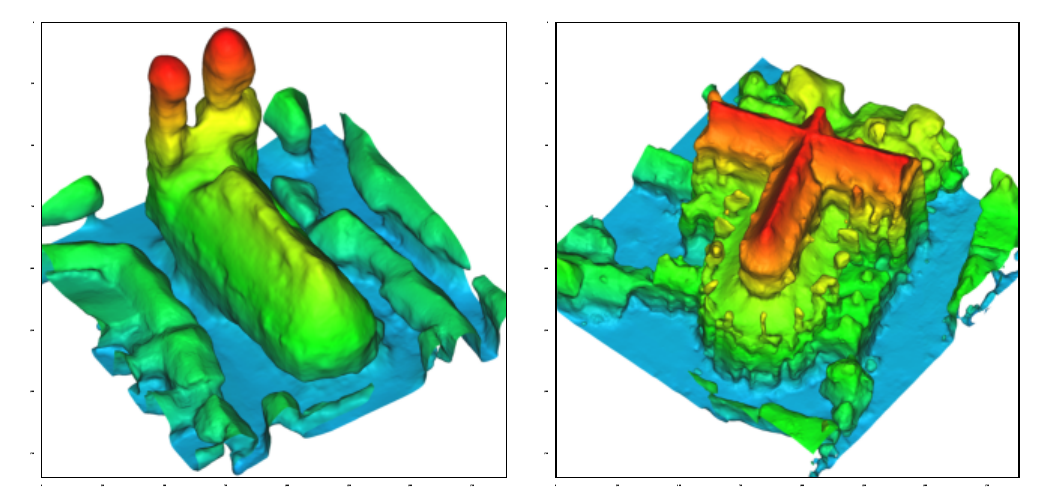}
	\caption{Intermediate reconstructions based on depth and smoothness supervision used as geometric initialization for the main training stage.}
\label{fig:inits}
\end{center}
\end{figure}

\sloppy
In a warm-up stage we train the network with depth supervision and smoothness regularization only. These models (figure \ref{fig:inits}) serve as geometric initialization for the main training stage with photometric supervision enabled. The warm-up optimization typically converges within 1\,k epochs, which equals approximately 6 minutes of training on our hardware. We found the parameters in table \ref{tab:training_parameters} to be a good balance between detail provided by tie points and completeness of reconstructed surfaces. Despite the sparsity of tie points, completeness of reconstructions is rather impressive. 

\begin{figure}[ht!]
\begin{center}
		\includegraphics[width=1.0\columnwidth]{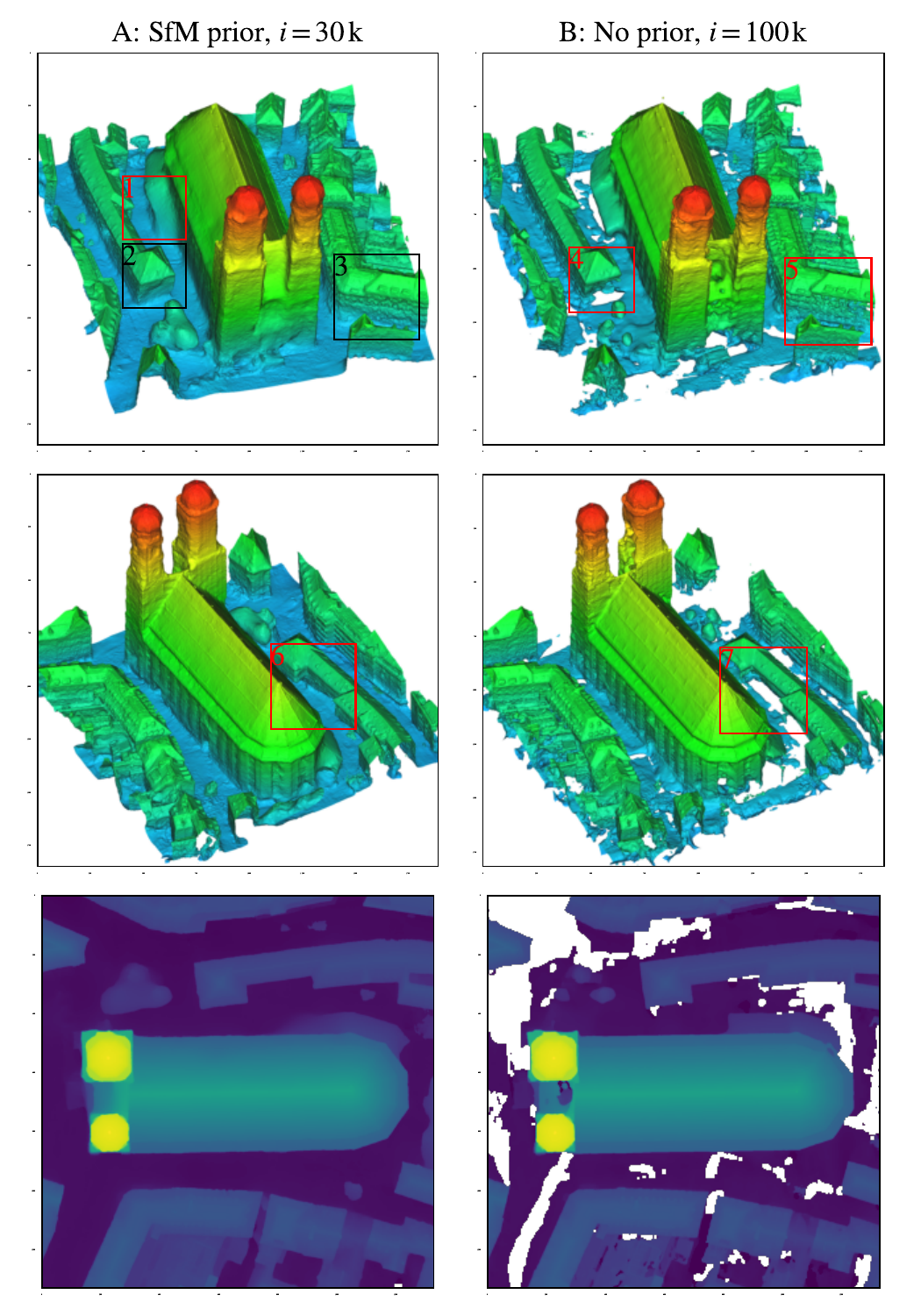}
	\caption{Results after 30\,k (left) and 100\,k (right) training epochs for the \textit{Frauenkirche} dataset using the sparse, depth priors from tie points (left) or only RGB input (right) for training. The first two rows display extracted meshes from different viewpoints. Row three shows extracted DSMs.}
\label{fig:muc_tpcont_imgs}
\end{center}
\end{figure}

\begin{figure}[ht!]
\begin{center}
		\includegraphics[width=1.0\columnwidth]{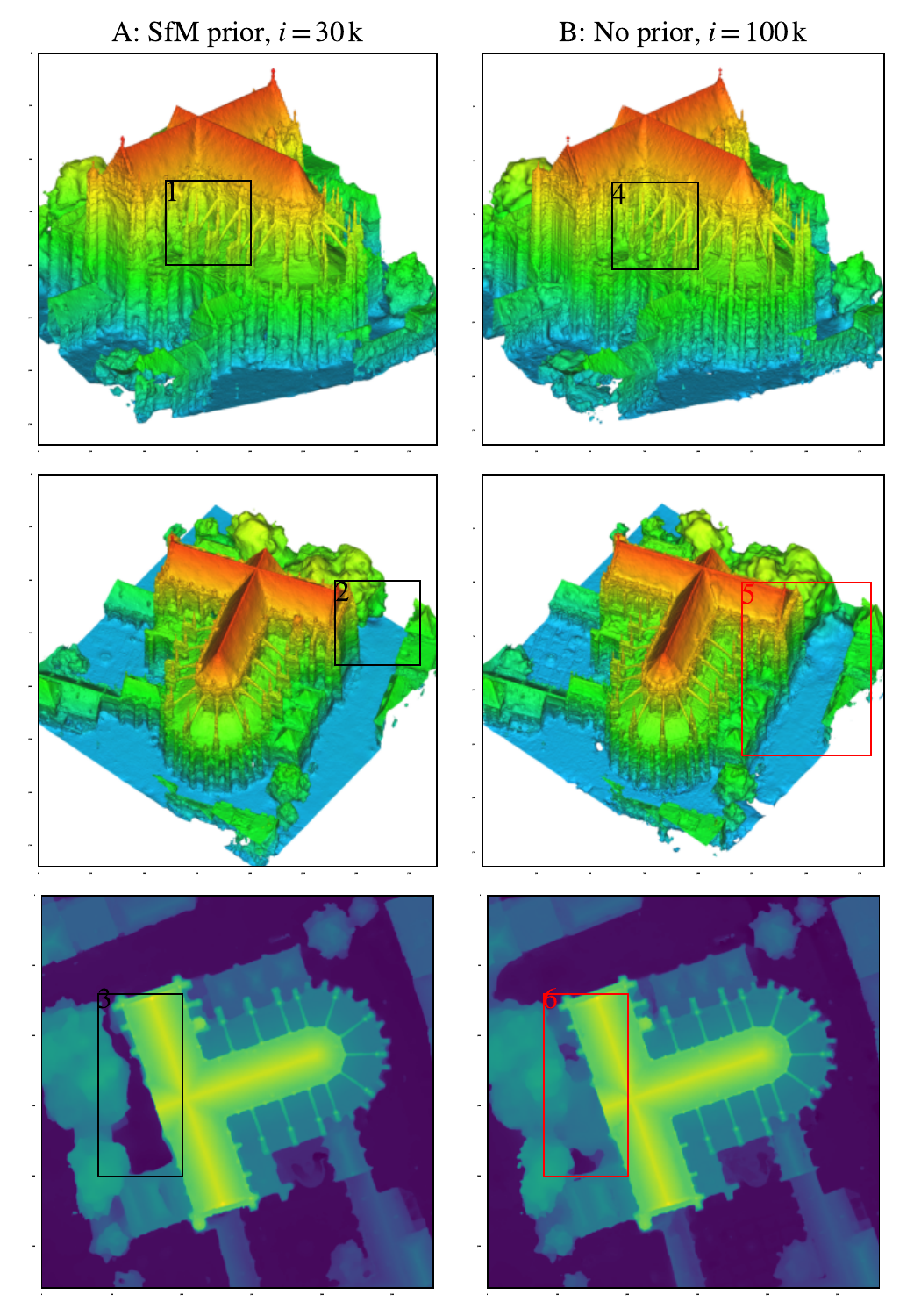}
	\caption{Results after 30\,k (left) and 100\,k (right) training epochs for the \textit{Domkirk} dataset using the sparse, depth priors from tie points (left) or only RGB input (right) for training. The first two rows display extracted meshes from different viewpoints. Row three shows extracted DSMs.}
\label{fig:dom_tpcont_imgs}
\end{center}
\end{figure}

\begin{figure}[ht!]
\begin{center}
		\includegraphics[width=1.0\columnwidth]{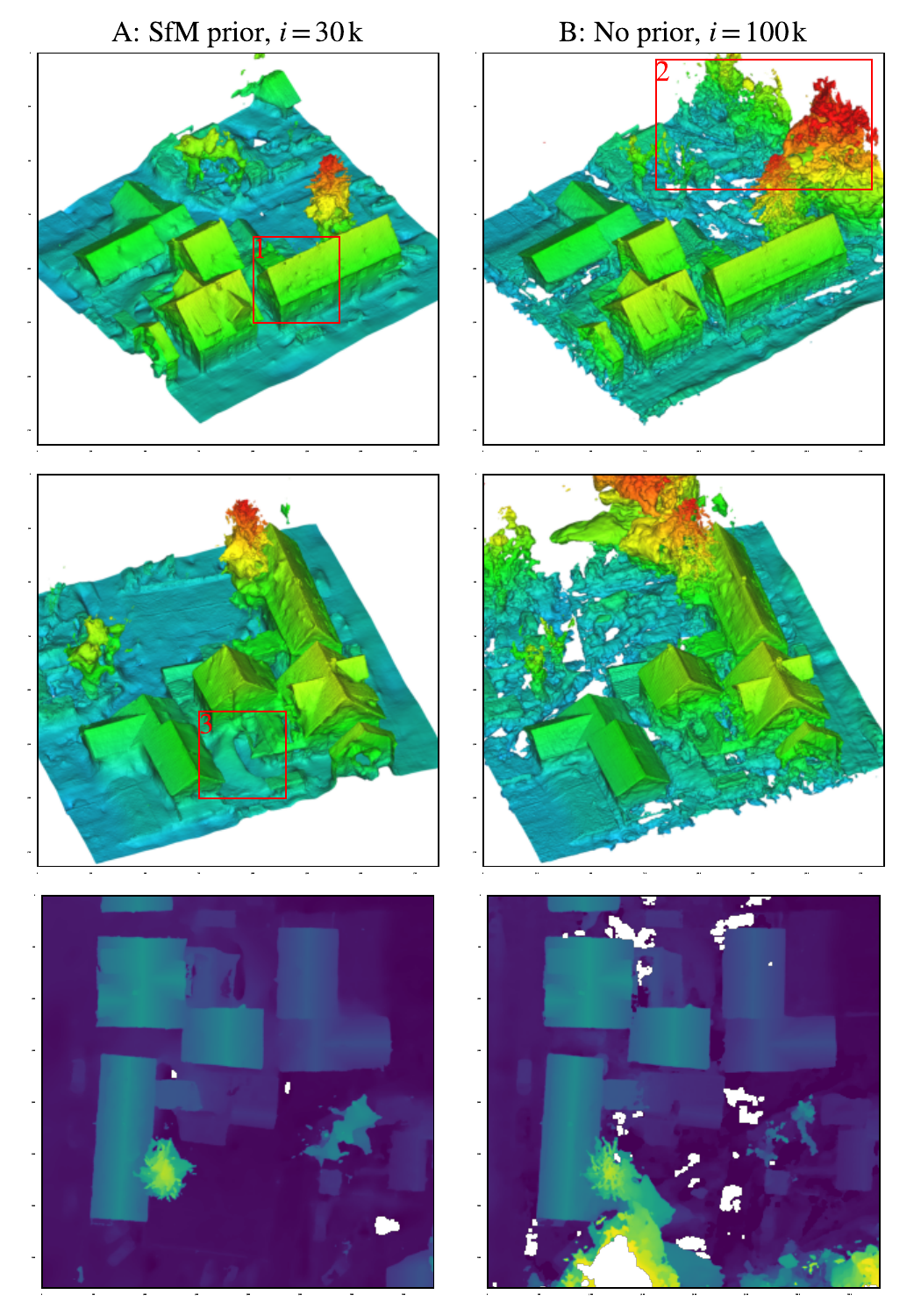}
	\caption{ Results after 30\,k (left) and 100\,k (right) training epochs for the \textit{Hessigheim 3D} dataset using the sparse, depth priors from tie points (left) or only RGB input (right) for training. The first two rows display extracted meshes from different viewpoints. Row three shows extracted DSMs.}
\label{fig:hd3_tpcont_imgs}
\end{center}
\end{figure}

Figures \ref{fig:muc_tpcont_imgs}, \ref{fig:dom_tpcont_imgs} and \ref{fig:hd3_tpcont_imgs} display the results for VolSDF with depth prior after 30\,k epochs (first column) and vanilla VolSDF after 100\,k epochs (second column) for all evaluated datasets. We found that additional training improves details but does not fix erroneous topology of extracted surfaces. The boxes in figures \ref{fig:muc_tpcont_imgs}, \ref{fig:dom_tpcont_imgs} and \ref{fig:hd3_tpcont_imgs} highlight areas of sub-optimal reconstruction (red) and improvements (black) achieved by using depth supervised VolSDF. 

Without depth prior VolSDF has difficulties reconstructing complete surfaces when trained on \textit{Frauenkirche}. The ground level is incomplete, presumably due to (moving) shadows and weak texture. Furthermore, VolSDF fails in the reconstruction of facades (figure \ref{fig:muc_tpcont_imgs}, boxes 4, 5 and 7). We assume that this is caused by the combination of limited number of observations and repetitive structure.
Topology significantly improves when using depth-supervised VolSDF. The geometric initialization ensures more complete ground surface. Despite the limited number of tie points on facades, depth supervision improves completeness (figure \ref{fig:muc_tpcont_imgs}, boxes 2 and 3). In areas where no tie points are computed, no improvements can be observed (box 6).

For the \textit{Domkirk} dataset impressive detail is reconstructed for both approaches (figure \ref{fig:dom_tpcont_imgs}, box 1 and 2).  Again VolSDF gets stuck in local minima in weakly observed and low-texture areas (box 6 and 5). In both cases depth supervision facilitates reconstruction of a more correct surface (box 2 and 3).  

Similar to \textit{Frauenkirche} depth supervised VolSDF delivers more complete reconstructions for the \textit{Hessigheim 3D} scene. Furthermore, VolSDF struggles to reconstruct areas for which appearance across views is dissimilar or contradictory, e.g vegetation (figure \ref{fig:hd3_tpcont_imgs}, box 2). For such areas the depth prior resolves ambiguities and constrains the optimization resulting in more faithful surfaces. We note our approach seems rather robust to imprecise or outlier contaminated tie points and only in rare cases generates artifacts as spikes (box 1).

 

\subsection{Quantitative Evaluation}
The number of publically available benchmarks for aerial surface reconstruction is very limited. The main challenge is to collect precise and geo-referenced 3D ground truth data featuring sufficient density to evaluate reconstruction quality of sharp depth discontinuities and small details. We base our quantitative evaluation on DSM raster data of the \textit{Frauenkirche} scene. As ground truth we use the benchmark DSM provided by \cite{Haala2013TheLO}. This DSM was generated by robust fusion of eight DSMs computed by eight independent reconstruction pipelines across academia and industry. 

\paragraph{Error metrics.} We evaluate the correctness of the reconstruction in terms of \textit{accuracy} and \textit{completeness} as defined in the ETH3D benchmark \cite{schoeps2017cvpr}. For both metrics, we evaluate differences between corresponding height values of ground truth DSM and a DSM derived from our reconstruction. For the latter we generate point clouds on the SDF zero level set and rasterize the highest surface points in the area of interest. Both metrics are evaluated over a range of tolerance thresholds, ranging from 1\,GSD to 30\,GSD. 





Additionally, we measure the noise of our surfaces in a robust fashion. The \textit{NMAD} metric measures the similarity of the prediction and the ground truth within the error band, i.e. is an indicator of the amount of noise within the tolerance. More specifically, NMAD is the Normalized Median Absolute Deviation and is a robust estimator for the standard deviation in normally distributed data \cite{nmad}.



\begin{figure}[ht!]
\begin{center}
		\includegraphics[width=1.0\columnwidth]{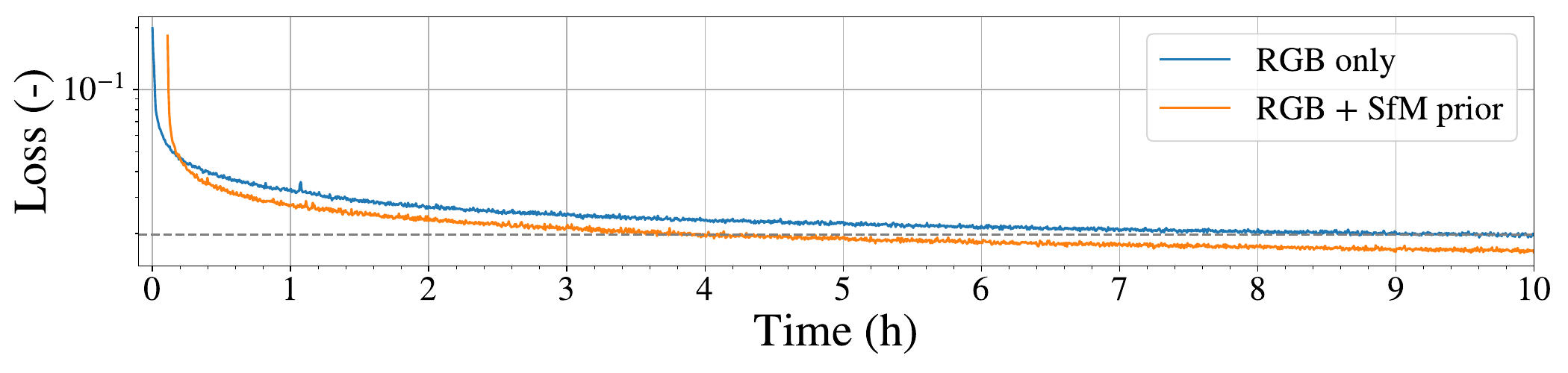}
	\caption{Loss curves for training based on \textit{Frauenkirche} data using RGB input only (blue), or additional depth priors (orange) from SfM/AT in form of tie points (TPs). The grey, dashed line shows the loss after 10\,hours of training when using RGB only.}
\label{fig:loss_curves_sfmrgb_muc}
\end{center}
\end{figure}

\begin{figure}[ht!]
\begin{center}
		\includegraphics[width=1.0\columnwidth]{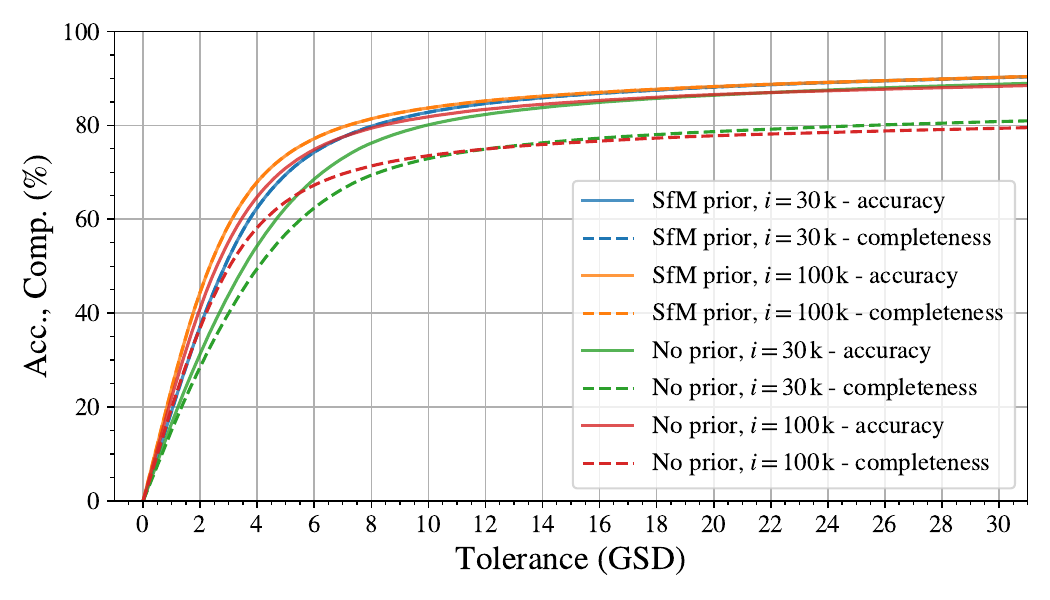}
	\caption{\textit{Accuracy} (solid line, left) and \textit{completeness} (dashed line, left) scores for the \textit{Frauenkirche} dataset using the sparse, depth priors from tie points for training.}
\label{fig:muc_tpcont_metrics_acc_comp}
\end{center}
\end{figure}

\begin{figure}[ht!]
\begin{center}
		\includegraphics[width=1.0\columnwidth]{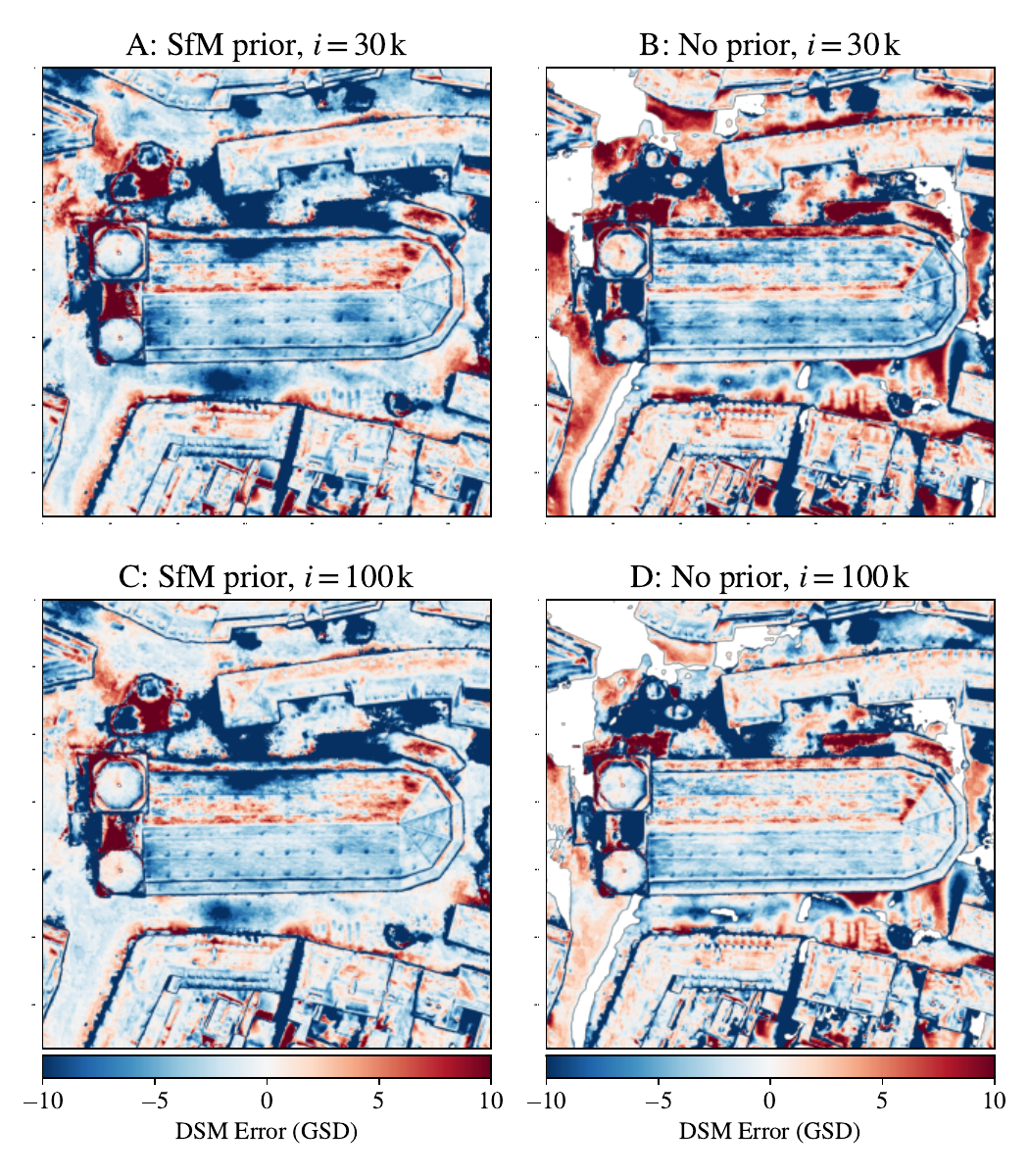}
	\caption{Differences between the extracted DSMs and the GT DSM, where the colors saturate outside of ±10 GSD.}
\label{fig:muc_tpcont_dsm_diff}
\end{center}
\end{figure}

\begin{figure}[ht!]
\begin{center}
		\includegraphics[width=1.0\columnwidth]{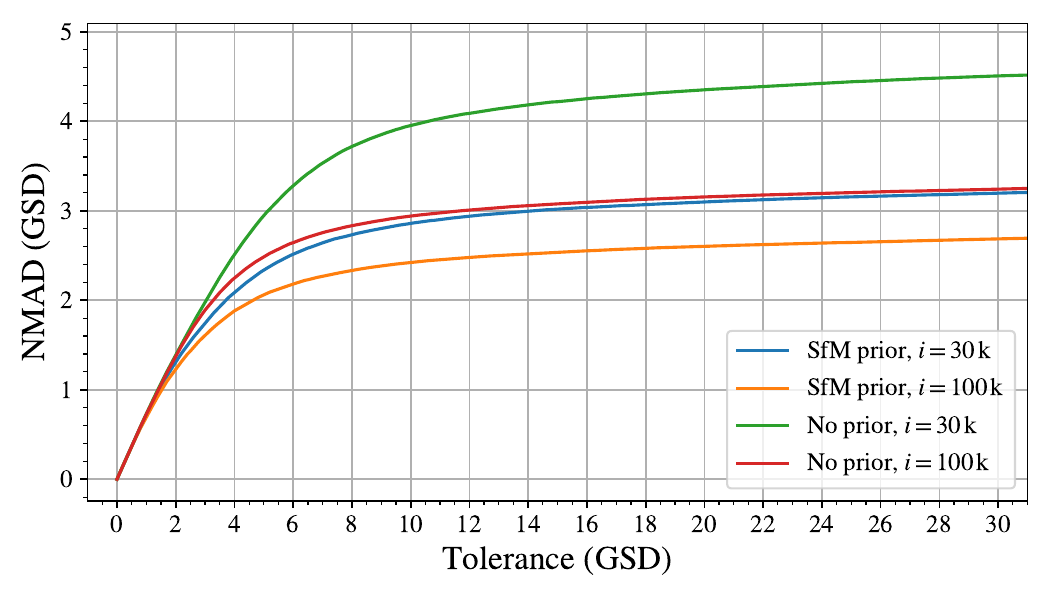}
	\caption{\textit{NMAD} scores for the \textit{Frauenkirche} dataset using the sparse, depth priors from tie points for training. }
\label{fig:muc_tpcont_metrics_nmad}
\end{center}
\end{figure}

\paragraph{Reconstruction Quality.}
Figures \ref{fig:loss_curves_sfmrgb_muc}, \ref{fig:muc_tpcont_metrics_acc_comp}, \ref{fig:muc_tpcont_dsm_diff} and \ref{fig:muc_tpcont_metrics_nmad} display metric scores of the models trained on \textit{Frauenkirche} data. We show scores for VolSDF and the depth-supervised variant after 30\,k and 100\,k epochs. 

Figure \ref{fig:muc_tpcont_metrics_acc_comp} displays the \textit{accuracy} and \textit{completeness} scores. In terms of \textit{completeness}, our model trained for 30\, k epochs significantly outperforms the reference model trained for 100\,k epochs. The respective scores converge with a difference of around 10\,\%. This validates the observation that local minima and incompleteness are mitigated in early training stages already.
The reconstruction \textit{accuracy} of our model is higher than that of the reference model throughout the entire tolerance interval when both are evaluated after 30\,k epochs, verifying accelerated convergence. Furthermore, our model after 30\,k iterations is almost on par with the reference trained for 100\,k epochs, which only delivers a slightly better score for lower GSD ranges. Our model trained for 100\,k epochs achieves the best \textit{accuracy} throughout the entire evaluation range. 
Figure \ref{fig:muc_tpcont_dsm_diff} visualizes the signed differences between the reconstructed and ground truth DSMs for VolSDF with depth prior (left column) to vanilla SDF (right column). 
After 30\,k epochs of training we observe improved quality of the depth supervised variant (A and B), in particular for the streets around the building. Furthermore, additional training further improves quality of both solutions although the progress is rather slow. We note that very inaccurate areas are not improved even beyond 100\,k epochs.

Accelerated convergence for depth supervision can also be observed in the loss curves over training time (figure \ref{fig:loss_curves_sfmrgb_muc}). Right after initialization, the loss of the depth-guided model drops significantly faster compared to the baseline. After the first hour of training, however, the values of both curves are only slowly decreasing. 
This underlines the observation that in early training stages the depth priors rapidly guide the reconstruction to a faithfull solution. In later training stages, details are refined which for both approaches still demands considerable computation time.   


Depth-supervised VolSDF  outperforms vanilla VolSDF in terms of \textit{NMAD} scores (figure \ref{fig:muc_tpcont_metrics_nmad}). Notably even after 30\,k iteations the \textit{NMAD} is slightly better than training its counterpart for 100\,k iterations. After 100\,k iterations we achieve a \textit{NMAD} score below 3\,GSD. 

\paragraph{Processing Time.} The use of tie point priors improves convergence in the early training stages, thus decreases runtimes. We note that it is difficult to rigorously compare training times across the approaches since surface states as well as final solutions differ. However, for the majority of scene parts comparable visual reconstruction quality of vanilla VolSDF and its depth supervised variant is obtained after 100k iterations vs 30k iterations respectively. For both approaches, training for 30\,k and 100\,k iterations takes about 3:20\,h and 11\,h respectively.
 
\begin{figure}[ht!]
\begin{center}
		\includegraphics[clip, trim=10cm 3cm 0cm 3cm, width=0.75\columnwidth]{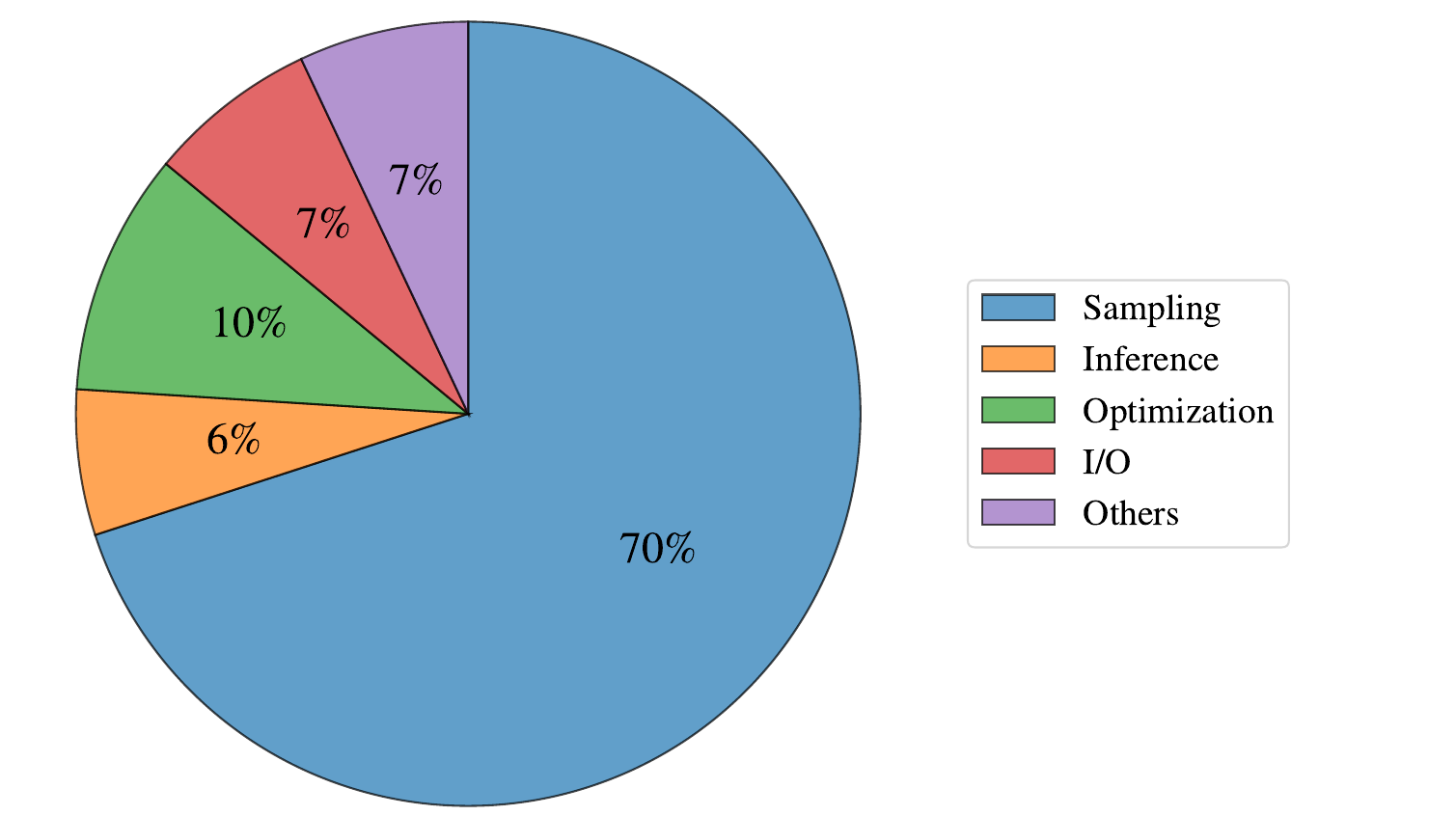}
	\caption{Relative time requirements of different model components within one epoch. The sampling algorithm accounts for 70\,\% of training time.}
\label{fig:profiling}
\end{center}
\end{figure}

Furthermore, we profiled the algorithm to identify most computationally expensive routines. Figure \ref{fig:profiling} shows the relative time requirements of the different model components within one training epoch. The additional depth-supervision terms do not generate noticeable computational overhead. The main bottleneck is VolSDF's sampling routine accounting for 70\,\% of training time, which suggests future optimization.

\section {Conclusion}\label{Conclusion}
\sloppy

We present the applicability of VolSDF, a NeRF variant modeling implicit neural surfaces, for 3D reconstruction from airborne imagery. We demonstrated that supervising VolSFD by tie points improves reconstructions: we observed faster convergence in early training stages and better quality in terms of completeness and accuracy. This is in particular true for challenging areas featuring only limited data evidence for which VolSDF tends to get stuck in local minima or does not converge at all. Reconstructed surfaces of an example nadir scene featured less than 4 GSD deviations to traditional MVS pipelines in terms of \textit{NMAD}. To completely converge and recover full detail prolonged training times are still required. This hampers practical application. However, we obtain topologically correct surfaces in reasonable time which could be subject to subsequent mesh post-processing. Sampling routines are the main bottle neck in the evaluated implementation and subject to future work. On the one hand efficient GPU implementation could speed up this process \cite{neus2}, on the other hand we want to investigate possibilities to dynamically reinforce sampling in areas with a large potential for improvements \cite{3DGaussianSplatting}. Neural implicit surface reconstruction is still an active research topic and we hope that this article encourages future work also in the domain of geometric reconstruction from aerial imagery and other remote sensing applications.

{
	\begin{spacing}{1.0}
		\normalsize
  \bibliography{Depth_Supervised} 
	\end{spacing}
}

\end{document}